\documentclass{article}

\usepackage{microtype}
\usepackage{graphicx}
\usepackage{subfigure}
\usepackage{booktabs} 
\usepackage{amssymb}
\usepackage{mathrsfs}

\usepackage{hyperref}

\usepackage[accepted]{icml2021}

\icmltitlerunning{Image Augmentation Based Momentum Memory Intrinsic Reward for Sparse Reward Visual Scenes}

\begin{document}

\twocolumn[
\icmltitle{Image Augmentation Based Momentum Memory Intrinsic Reward for\\
             Sparse Reward Visual Scenes}



\icmlsetsymbol{equal}{*}

\begin{icmlauthorlist}
\icmlauthor{Zheng Fang}{to}
\icmlauthor{Biao Zhao}{to}
\icmlauthor{Guizhong Liu}{to}
\end{icmlauthorlist}

\icmlaffiliation{to}{School of Information and Communications, Xi’an Jiaotong University, Xi’an, China}

\icmlcorrespondingauthor{Guizhong Liu}{liugz@xjtu.edu.cn}

\icmlkeywords{Machine Learning, ICML}

\vskip 0.3in
]



\printAffiliationsAndNotice{\icmlEqualContribution} 

\begin{abstract}
Many scenes in real life can be abstracted to the sparse reward visual scenes, where it is difficult for an agent to tackle the task under the condition of only accepting images and sparse rewards. We propose to decompose this problem into two sub-problems: the visual representation and the sparse reward. To address them, a novel framework \textbf{IAMMIR} combining the self-supervised representation learning with the intrinsic motivation is presented. For visual representation, a representation driven by a combination of the image-augmented forward dynamics and the reward is acquired. For sparse rewards, a new type of intrinsic reward is designed, the Momentum Memory Intrinsic Reward (\textbf{MMIR}). It utilizes the difference of the outputs from the current model ($online$ network) and the historical model ($target$ network) to present the agent's state familiarity. Our method is evaluated on the visual navigation task with sparse rewards in Vizdoom. Experiments demonstrate that our method achieves the state-of-the-art performance in sample efficiency, at least 2 times faster than the existing methods reaching 100\% success rate. 
\end{abstract}

\section{Introduction}

Deep reinforcement learning has grown by leaps and bounds in recent years, achieving superhuman performance in many video games \cite{vinyals2019grandmaster, berner2019dota, badia2020agent57}. However, when we desire to apply deep reinforcement learning to the real-world decision-making control tasks, like navigation \cite{tai2017virtual, zhelo2018curiosity} or manipulation \cite{sermanet2018time}, it has not achieved the similar astonishing performance like in games. We believe that an important reason for this is that the direct models of many real tasks are sparse reward visual scenes, that is, the observation of the agent is the high-dimensional image representation, and the feedback of the task only gives a positive reward when the agent completes the task and the rewards are 0 at the rest times. The RL agent would have no clue about what task to accomplish until it receives the terminal reward for the first time by chance. To fully figure out this problem, we propose to decompose it into two sub-problems: the visual representation and the sparse reward, which obstruct the agent to get efficiently trained.

First, the visual representation in DRL is quite significant when the agent has to make decision in visual controlled scenes. A promising approach is to learn a latent representation together with the control policy. Prior works \cite{srinivas2020curl, schwarzer2020data, kostrikov2020image, yarats2021reinforcement, laskin2020reinforcement} have shown that an auxiliary task, like self-supervised representation learning, with the standard image-based RL, leads to more robust and effective representations. However, such techniques are not so effective in settings where the environmental rewards are too sparse since fitting a high-capacity encoder needs diverse data and dense reward feedback.
 
Second, the sparse reward remains a hard problem in RL. We believe the main reason is that sparse reward cannot help the agent fully explore the environment. A common approach to exploration is the intrinsic motivation\cite{oudeyer2007intrinsic, schmidhuber2010formal} by generating intrinsic rewards. Existing formulations of intrinsic rewards include maximizing “visit count” of less-frequently visited states \cite{bellemare2016unifying, ostrovski2017count} , “curiosity” where the prediction error is used as the reward signal \cite{pathak2017curiosity, burda2018large}. In general, there can be mainly two concernings when dealing with the intrinsic motivations. One is the stochastic property of the agent-environment system, which made it quite unable to predict properly with a simple prediction model. The other is the state representation\cite{ yarats2021reinforcement}. Due to the high demensional property and irrelevant containts of the images that the intrisic reward can hardly be defined. Thus, a representation which can precisely capture the significant latent of the environment state is needed in order to distinguish the novel states from the visited ones.

We find that, both the visual representation and the sparse rewards are interrelated and valuable to the sparse reward visual scenes. Regarding to the control problems, the reward driven representation is critical, although the visual representation can be tackled by auxiliary tasks. While the sparse reward can be addressed by the intrinsic motivation, a good choice of the latent can make the computation of the intrinsic reward tractable and filter out the irrelevant aspects of the observations.

Focusing on the visual representation and the sparse reward, we have proposed the \textbf{IAMMIR} framework, which efficiently fuses the self-supervised representation learning with the intrinsic motivation. In summary, this paper makes the following contributions:
\begin{itemize}
\item To the visual representation, we propose a representation jointly driven by the image-augmented forward dynamics and reward. To the image-augmented forward dynamics driven, the image-augmented states are operated to make forward dynamics predictions, where the temporal features and the consistency can be well extracted. To the reward driven, we exploit a new RL objective that leverages the intrinsic and extrinsic rewards to steer the course of  the representation learning. Compared with the previous works \cite{pathak2017curiosity,srinivas2020curl}, the learned latent is better suits the control tasks. 
\item To the sparse reward, we propose a novel intrinsic reward \textbf{MMIR}, which utilizes the output error between the $online$ network and the $target$ network at the same state to present the agent's state familiarity. \textbf{MMIR} is unrelated to state transition, avoiding the interference of environment stochasticity, and its computation is based on the effective latent obtained from the visual representation module. In this way, intrinsic reward and self-supervised representation learning are efficiently coupled.
\item We demonstrate the ability of our method \textbf{IAMMIR} in tackling the sparse reward visual scenes, like the visual navigation tasks in Vizdoom, at least 2 times faster than the current state-of-the-art methods in reaching 100\% success rate, which indicates a high exploration efficiency and the state-of-the-art performance. Besides, we also experimentally demonstrate a certain scene generalization with our method.
\end{itemize}

The paper is organized as follows. In Section 2, we introduce the related work. In Section 3, we describe the problem model and our method in detail. In Section 4, experimental results on the visual navigation tasks are shown in comparison with the prior works. Finally, a conclusion is drown in Section 5.

\section{Related Work}

In this section, we provide a brief description on the most relevant work that our work builds on.

\textbf{Self-Supervised Learning} \quad In recent years, self-supervised learning has achieved great success in many fields, which extracts training signals from a large amount of unlabeled data and learns good representations to downstream tasks. There are mainly two types of methods in SSL, the self-predictive learning and the contrastive learning. Self-predictive learning \cite{vondrick2018tracking, wang2019learning} refers to the paradigm in which a model learns the ability to predict a portion of input from the remaining. Contrastive learning \cite{chen2020simple} is to learn a representation space where positive samples are close and negative samples are far apart. Self-supervised learning based on contrastive learning provides a strong initialization for the downstream tasks such as image classification \cite{chen2020simple, he2020momentum, grill2020bootstrap, caron2020unsupervised}. Our work is partly inspired by the method BYOL\cite{grill2020bootstrap}. We apply the same contrastive learning loss, but making corrections to the input and the network architecture to learn the temporal features.

\textbf{Visual Representation in RL} \quad
In the visual decision-making scenarios, learning a good representation can improve not only the sample efficiency but also the decision-making performance. There has been a lot of works in the visual decision-making scenarios, integrating reinforcement learning with self-supervised learning to acquire the visual representation. CURL\cite{srinivas2020curl} and SPR \cite{schwarzer2020data} encourage the discovery of consistent features with self-supervised loss function. More recently, image augmentations have shown significant success in learning representations \cite{kostrikov2020image, laskin2020reinforcement}. Our work is also partly inspired by SPR \cite{schwarzer2020data}. However, we adopt a easy forward prediction head to learn temporal features instead of a transition model. In addition to applying visual representation to policy learning, our work also applies it to the computation of the intrinsic reward.

\begin{figure*}[ht]
    \centering
    \includegraphics[scale=0.55]{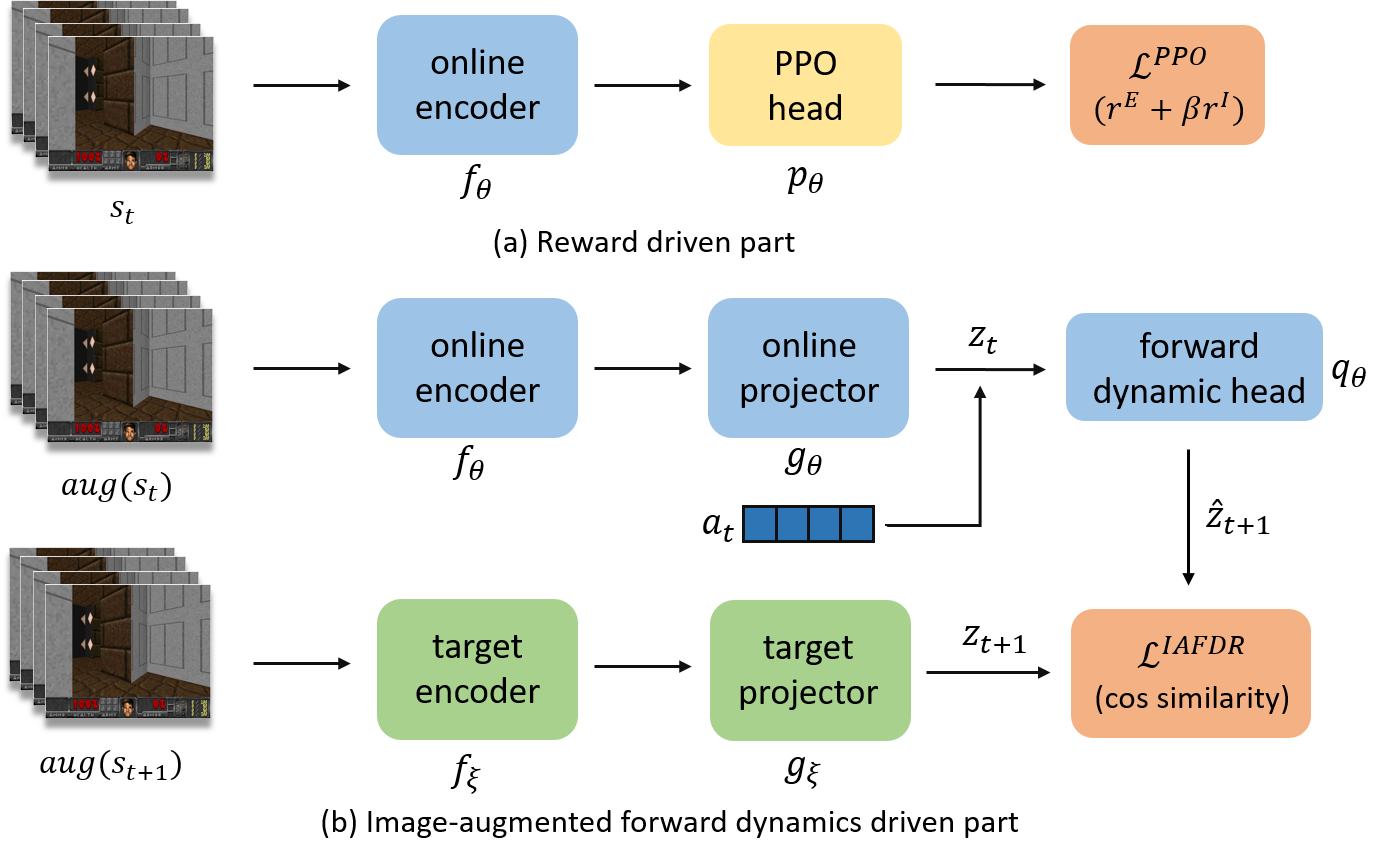}
    \caption{we propose a representation jointly driven by the image-augmented forward dynamics and the reward. In the reward driven part, the new RL objective consisting of the extrinsic and the intrinsic rewards drives the representation learning. In the image-augmented forward dynamics driven part, the image-augmented $aug(s_t)$ and $aug(s_{t+1})$ are projected into low-dimensional representations $z_t$ and $z_{t+1}$ through the encoders $f_\theta$ and $f_\xi$ and the projectors $g_\theta$ and $g_\xi$, and the forward dynamic head $q_\theta$ predicts $\hat{z}_{t+1}$ with $(z_t, a_t)$. The $\mathcal{L}^{IAFDR}$ cosine similarity loss function is used to acquire the representation. For details, see Section \ref{subsection:3.2}.}
    \label{fig1}
\end{figure*}

\textbf{Sparse Reward and Intrinsic Motivation in RL} \quad
As mentioned above, there are lots of control problems with the sparse reward setting in RL. The key is to fully explore the environment. Approaches that tackle this problem are generally task-agnostic. They exploit various inductive biases that correlate positively with the efficient exploration. Prior works include state visitation counts \cite{bellemare2016unifying, ostrovski2017count}, curiosity-driven exploration \cite{pathak2017curiosity, burda2018large}, distilling random networks \cite{burda2018exploration}, ensemble disagreement \cite{pathak2019self}, state reachability in episodic memory \cite{savinov2018episodic} and so on. Not as in the previous works, we propose a novel intrinsic reward \textbf{MMIR} to present the agent's state familiarity, which is quite simple and effective to encourage the agent to fully explore the environment.

\section{Method}
\label{section:3}
We consider the sparse reward visual scene as the Partially Observable Markov Decision Process (POMDP) with sparse reward setting, denoted as $(\mathcal{O}, \mathcal{A}, \mathcal{T}, \mathcal{R}, \gamma)$, where $\mathcal{O}$ represents a high-dimensional observation space, like image pixels, $\mathcal{A}$ is the action space, $\mathcal{T}:\mathcal{O} \times \mathcal{A} \rightarrow \mathcal{P}(\mathcal{O})$ is the observation transition probability, that is, the probability of the next observation given the last observation and action, $\mathcal{R}: \mathcal{O} \times \mathcal{A} \times \mathcal{O} \rightarrow \mathbb{R}$ is the temporal reward which is defined by an observation $o$, an action $a$ and the next observation $o’$. In our scenario, most of the rewards are 0, and only when the task is completed, a positive reward will be given. The parameter $\gamma \in [0,1]$ is the discount factor. To partially observable process, the general approach \cite{mnih2013playing} is to stack $k$ consecutive observations $\{o_{t-k},...,o_{t}\}$ to represent the state $s_t$, thereby converting the POMDP into a Markov decision problem\cite{bellman1957markovian} $(\mathcal{S}, \mathcal{A}, \mathcal{T}, \mathcal{R}, \gamma)$. 

We seek to train a policy $\pi(\cdot|s):\mathcal{S} \rightarrow \mathcal{P}(\mathcal{A})$ whose expected cumulative discounted rewards $\mathbb{E}_{\tau\sim\pi}[\sum_{t=0}^T\gamma^{t}r_{t}]$ is maximized in POMDP with sparse reward. To do this, we combine a strong model-free RL algorithm, PPO\cite{schulman2017proximal} with our \textbf{IAMMIR} as an auxiliary task to improve sample efficiency. In Section 3.1, PPO is briefly introduced. In Section 3.2 and 3.3, we describe the design of visual representation module and momentum memory intrinsic reward respectively in our \textbf{IAMMIR}. 

\subsection{Proximal Policy Optimization(PPO)}
The traditional policy gradient algorithm is the on-policy method in RL, where the behavior policy should be the same as the target policy. It just allows a sample to be used only once. In order to improve the sample efficiency, PPO utilizes the importance sampling to adjust the sample distribution. However, when the probability ratio $c_t(\theta) = \frac{\pi_\theta(a_t|s_t)}{\pi_{\theta_{old}}(a_t|s_t)}$ is far away from 1, the importance sampling would introduce a large variance and make the training process unstable. A simple clip on the probability ratio $c_t$ is adopted to regulate optimization update size in Eq (\ref{eq:1}). In the estimation of the advantage value function, GAE \cite{schulman2015high} is adapted in Eq (\ref{eq:2}), which also effectively reduces the variance of the gradient estimation.
\begin{equation}
\mathcal{L}^{PPO}(\theta) = \mathbb{E}_t[\min{(c_t(\theta)A_t, clip(c_t(\theta), 1-\epsilon, 1+\epsilon)A_t)}]
\label{eq:1}
\end{equation}
\begin{equation}
A_t = \sum_{k=t}^{T-1}(\gamma\lambda)^{k-t}\delta_{k}
\label{eq:2}
\end{equation}
\begin{equation}
\delta_{k} = r_k+{\gamma}V(s_{k+1}) - V(s_k)
\label{eq:3}
\end{equation}
where $clip$ is the operation to clip the value of $c_t(\theta)$ to $[1-\epsilon, 1+\epsilon]$ and $A_t$ is the advantage value function, estimated by the TD-error $\delta_{k}$.

\subsection{Forward Dynamic and Reward Driven Feature}
\label{subsection:3.2}
In the standard DRL, the representation of the state is driven by the reward, but in the sparse reward visual scenes, representation driven by a quite sparse reward could not fully describe the environment state. We propose to jointly train an image-augmented forward dynamics representation \textbf{IAFDR}. Specifically, on the basis of the forward dynamics model, the input image is processed by image augmentation in advance, which can 
extract the temporal feature and consistency simultaneously. Our visual representation module can be seen in Fig \ref{fig1}, which consists of the following four parts:

\textbf{Encoder} \quad
We use a multi-layer convolution network as the encoder $f$, as same as the architecture from ICM \cite{pathak2017curiosity}. Specifically, each state $s_t$ is processed by an image augmentation and the online encoder $f_\theta$ to obtain the representation $f_{\theta}(aug(s_t))$. At the next state $s_{t+1}$, the image augmentation is also executed. In order to prevent from the collapse of representation, the same model architecture target encoder $f_{\xi}$ is applied with its parameters updated by the exponential moving average (EMA). The EMA factor is $\tau \in [0,1]$.
\begin{equation}
\xi \leftarrow \tau\xi + (1-\tau)\theta
\label{eq:4}
\end{equation}

A random shift and brightness transformation as the image augmentation like in the existing works is utilized \cite{srinivas2020curl, schwarzer2020data}. Kornia \cite{riba2020kornia}, for instanve, is exploited for efficient GPU-based image augmentations.

\textbf{Projector} \quad
Similar to the network structure of the contrastive learning \cite{chen2020simple}, the projector $g$ is also used to compact the representation output by the encoder $f$ to a low-dimensional feature $z=g(f(aug(s)))$. There are  online and target projectors $g_{\theta}$ and $g_{\xi}$. The target projector parameters are given by an EMA of the online projector parameters, using the same update as the online and target encoders. Besides being refined more significant information in visual representation, the projector is also utilized as the state representation in the intrinsic reward computation to mitigate its ill effects in the new RL objective (more details in \ref{sbusection:4.4}).

\textbf{Forward Dynamic Head} \quad
After the encoder $f$ and the projector $g$, we get the state representations $(z_t, z_{t+1})$ at two consecutive times $t$ and $t+1$. A forward dynamics head $\hat{z}_{t+1} = q_{\theta}(z_t,a_t)$ is trained to predict the next state representation $z_{t+1}$, which plays an important role in avoiding the representation collapse in the contrastive learning. Finally, the loss function of the image-augmented forward dynamic representation \textbf{IAFDR} is:
\begin{equation}
\mathcal{L}^{IAFDR}_{\theta}(s_t,a_t,s_{t+1}) = -cos(\hat{z}_{t+1}, z_{t+1})
\label{eq:5}
\end{equation}

\textbf{Reward Driven Feature} \quad
In the sparse reward scenes, it is hardly possible to learn a good policy from the image by using the rewards from the environment. A fatal reason is that the sparse rewards cannot drive the encoder to learn a sufficient state representation, which in turn affects the policy learning. To acquire better reward driven representation, we design a new type of intrinsic reward \textbf{MMIR} aiming at exploring the environment efficiently. Through the combination of the intrinsic reward $r^I$ and the extrinsic reward $r^E$, a dense reward function $r^{I+E}=r^{E}+\beta r^{I}$ is generated, with factor $\beta$ reflecting the degree of exploration. The new reward function results in a new RL objective in Eq (\ref{eq:6}) and affects the TD-error $\delta_{k}$ in PPO:

\begin{equation}
\mathcal{J}^{new}(\theta) = \mathbb{E}_{\tau\sim\pi(\theta)}[\sum_{t=0}^T\gamma^{t}r_{t}^{I+E}]
\label{eq:6}
\end{equation}

\begin{equation}
\delta_{k} = r_{k}^{I+E}+{\gamma}V(s_{k+1}) - V(s_k)
\label{eq:7}
\end{equation}

In the training process, the total loss $\mathcal{L}^{total}(\theta)$ includes the image-augmented forward dynamics representation loss $\mathcal{L}^{IAFDR}(\theta)$ and the reinforcement learning loss $\mathcal{L}^{PPO}(\theta)$ in Eq (\ref{eq:8}). The image-augmented forward dynamics representation loss affects the encoder $f_{\theta}$, the projector $g_{\theta}$ and the forward dynamic head $q_{\theta}$. In PPO, the reinforcement learning loss affects the encoder $f_{\theta}$ and the ppo head $p_{\theta}$.
\begin{equation}
\mathcal{L}^{total}(\theta)=\mathcal{L}^{PPO}(\theta) + \alpha \mathcal{L}^{IAFDR}(\theta)
\label{eq:8}
\end{equation}

\begin{figure*}[ht]
    \centering
    \includegraphics[scale=0.6]{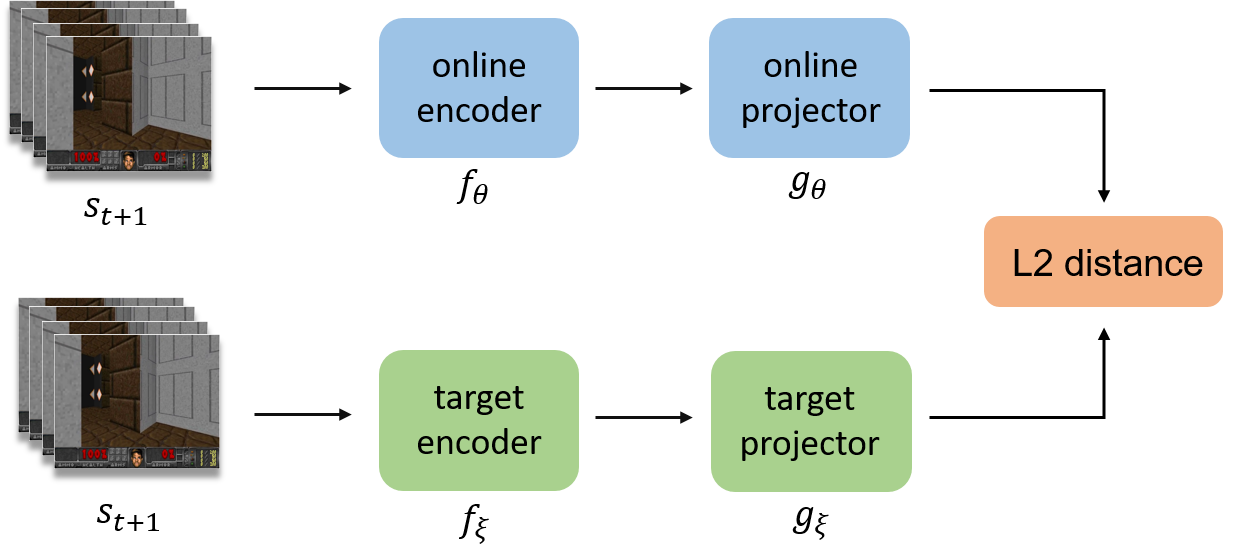}
    \caption{\textbf{MMIR}: We propose a new efficient and simple intrinsic reward which exploits the output error of two neural networks with momentum update. this design takes full advantage of representation learning while avoiding the uncertainty caused by state transition. The intrinsic reward $r^I(s_t, a_t, s_{t+1})$ is the L2 distance between the outputs of $s_{t+1}$ through the online network $(f_{\theta} \circ g_{\theta})$ and the target network $(f_{\xi} \circ g_{\xi})$. For details, see Section \ref{subsection:3.3}}
    \label{fig2}
\end{figure*}

\subsection{Momentum Memory Intrinsic Reward (\textbf{MMIR})}\label{subsection:3.3}
To encourage the agent to explore novel states, we design a new type of intrinsic reward to represent the agent's state familiarity, visualized in Fig \ref{fig2}. The smaller the intrinsic reward is, the more familiar the agent is with the state. We hypothesize that the difference of outputs from the current model and its historical model at the same sample can express the model's familiarity to the samples. Similarly, we use the output error of the two networks in the visual representation module to represent the agent's state familiarity in Eq (\ref{eq:9}). Specifically, the current model is composed of  the online encoder and the online projector $(f_{\theta} \circ g_{\theta})$, and the historical model is expressed by the target network, which consists of the target encoder and the target projector $(f_{\xi} \circ g_{\xi})$ with the momentum update. Momentum update, also known as EMA, can be understood as the temporal ensembling of the models with exponential weights. Hence, the target network can be seen as an ensemble of the online network’s current version and those earlier versions, and the EMA coefficient $\tau$ determines how many of the earlier versions mainly affects the target network output. For familiar states, the earlier models in the target network get the similar outputs as the online network. While, for novel states, the earlier models in the target network get different results from each other, which is far away from the online network. Since the intrinsic reward is computed by two networks whose parameters have a momentum relationship, we call it momentum memory intrinsic reward (\textbf{MMIR}).
\begin{equation}
r^I_t(s_t,a_t,s_{t+1})= ||g_{\theta}(f_{\theta}(s_{t+1}) - g_{\xi}(f_{\xi}(s_{t+1})||^2_2
\label{eq:9}
\end{equation}

In the sparse reward visual scenes, the challenge of intrinsic reward design is the uncertainty of the state transition and the meaningful state representation. As said in \cite{burda2018large}, the uncertainty of the state transition would make the prediction errors between time $t$ and time $t+1$ consistently high. In our method, however, \textbf{MMIR} only takes advantage of the output error at time $t+1$ of the two networks with momentum update as the intrinsic reward for time $t$, which avoids  the effects of stochastic transition. Meanwhile, the state representation is also well expressed by the representation jointly driven by the image-augmented forward dynamics and the reward.
\begin{figure}[h]
    \centering
    \subfigure[VizdoomMyWayHome]{
    \begin{minipage}[t]{1.\linewidth}
    \centering
    \includegraphics[scale=0.42]{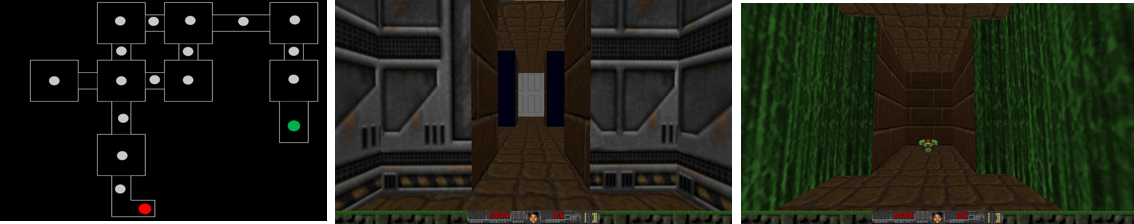}
    \end{minipage}%
    }%
    
    \subfigure[VizdoomFlytrap]{
    \begin{minipage}[t]{1.\linewidth}
    \centering
    \includegraphics[scale=0.42]{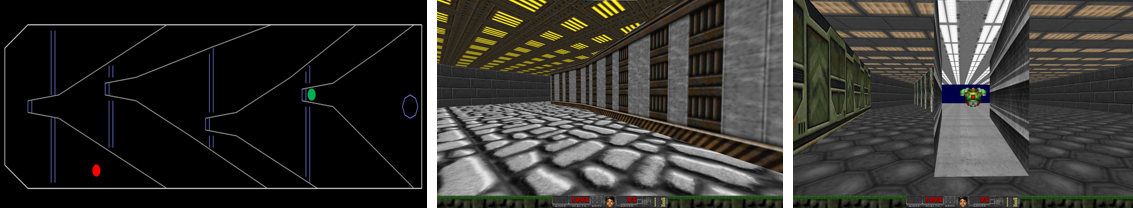}
    \end{minipage}%
    }%
    \caption{the top-down view, first-person view and the terminal state from the VizDoomMyWayHome and VizdoomFlytrap environment. In $VerySparse$, the agent start at the farthest position (red point) in (a). In $Dense$, the agent is randomly spawned in 17 locations (all points) in (a). In $Flytrap$, the agent start at the first room (red point) in (b). The agent needs to explore the environment until it finds the armour (green point) that triggers an extrinsic reward +1.
    }
    \label{fig3}
\end{figure}

\begin{figure*}[h]
    \centering
    \subfigure[$VerySparse$]{
    \begin{minipage}[t]{0.34\linewidth}
    \centering
    \includegraphics[scale=0.025]{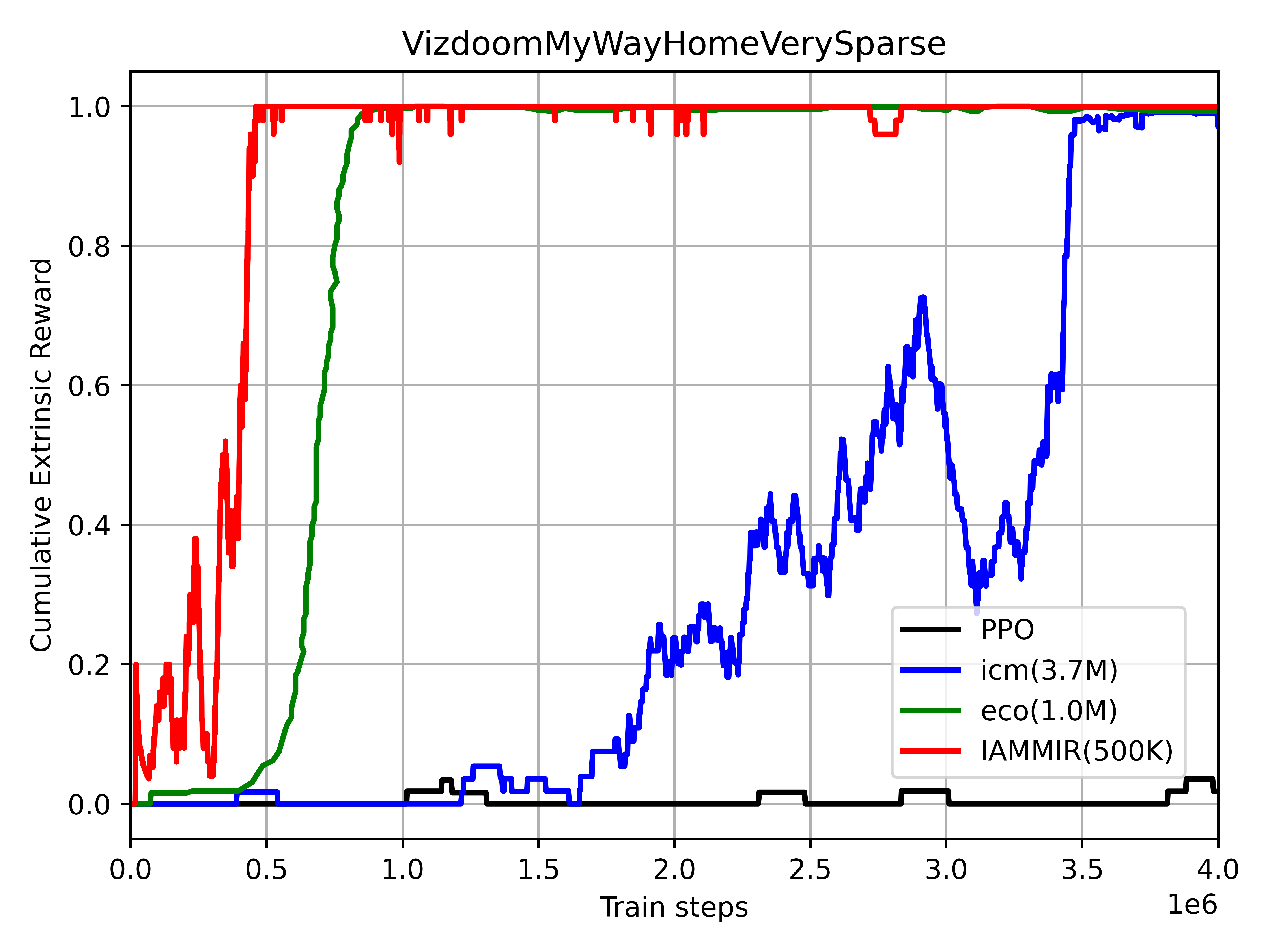}
    \end{minipage}%
    }%
    \subfigure[$Flytrap$]{
    \begin{minipage}[t]{0.33\linewidth}
    \centering
    \includegraphics[scale=0.025]{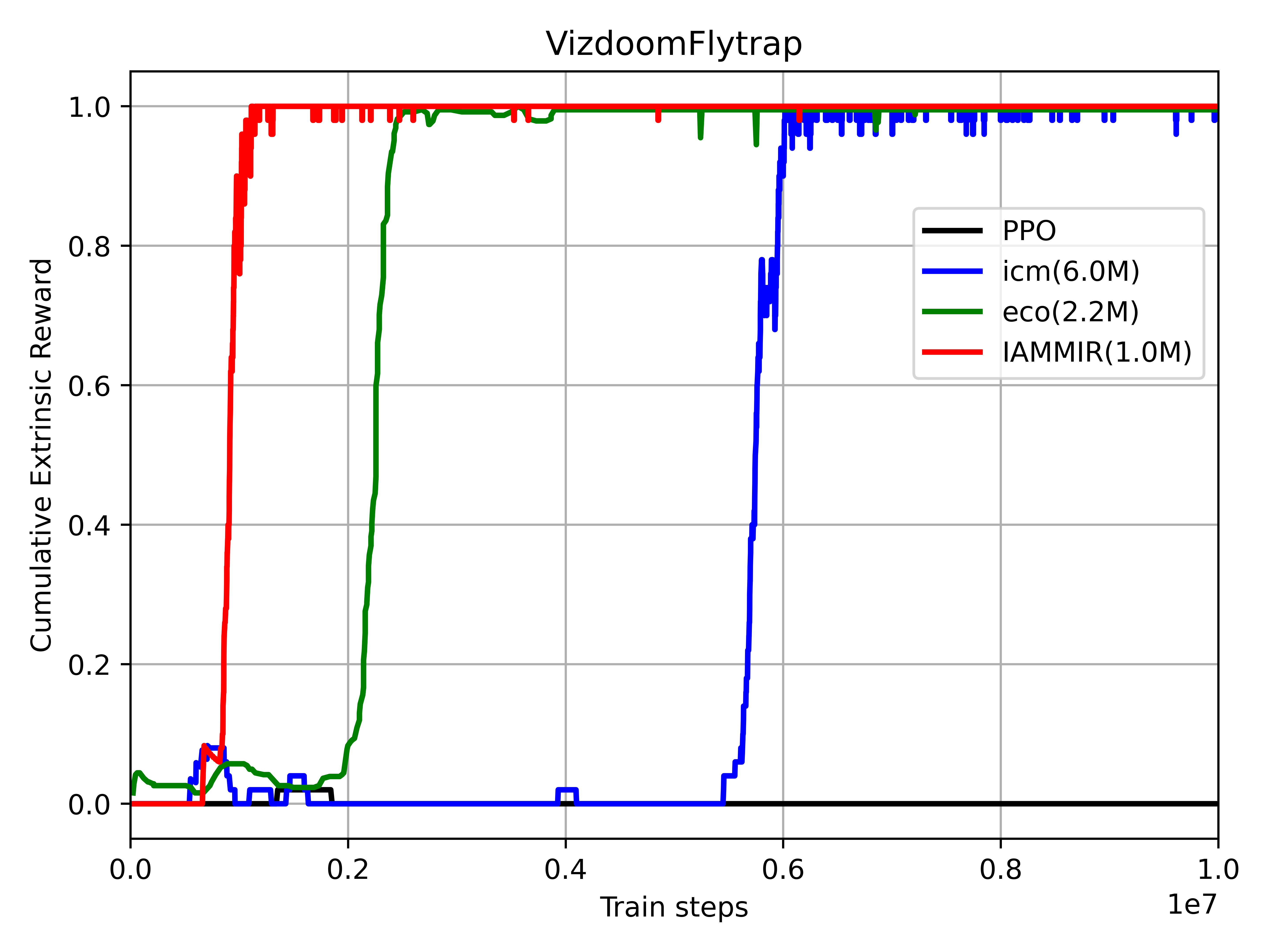}
    \end{minipage}%
    }%
    \subfigure[$Dense$]{
    \begin{minipage}[t]{0.33\linewidth}
    \centering
    \includegraphics[scale=0.025]{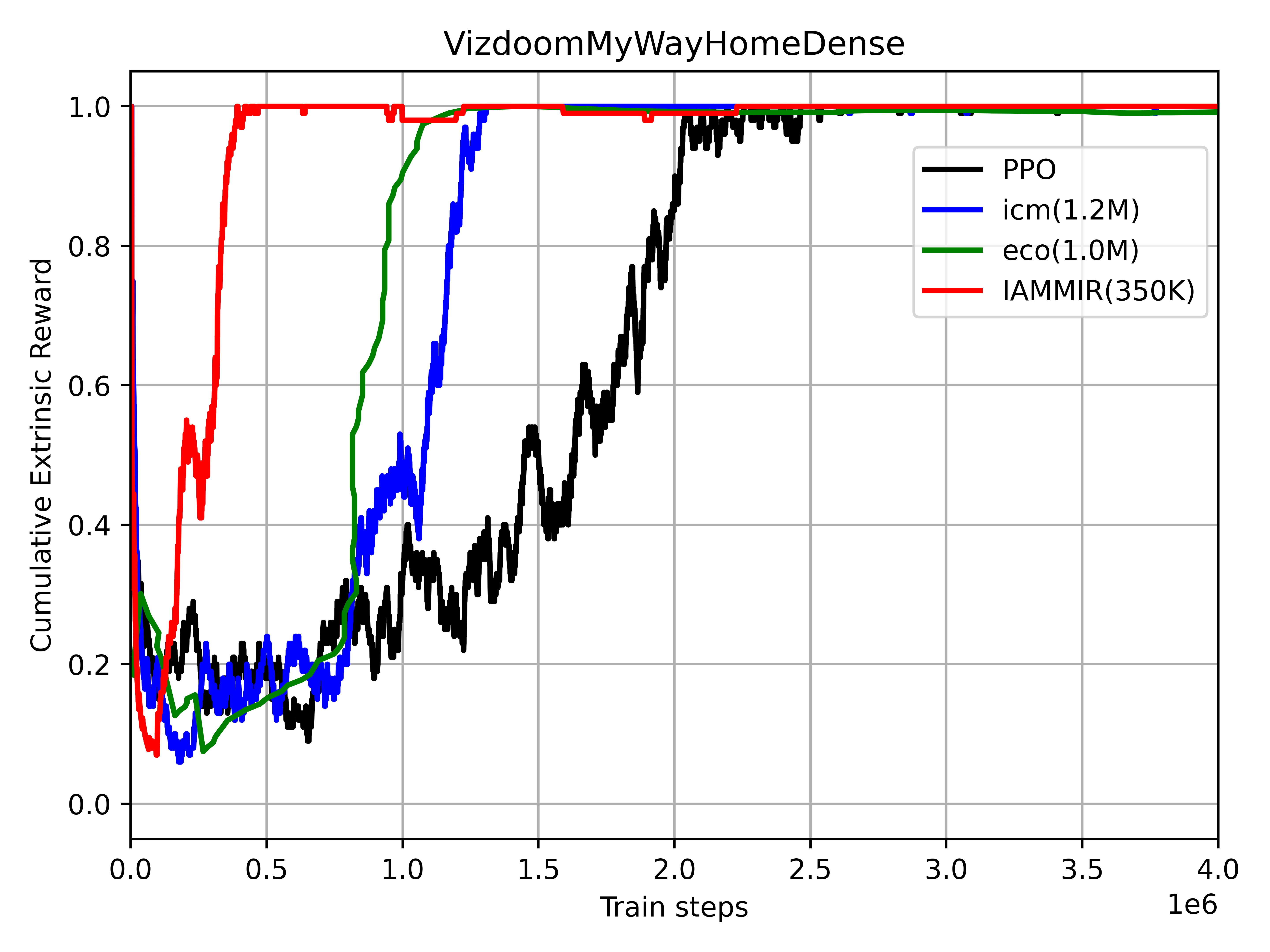}
    \end{minipage}%
    }%
    \caption{the cumulative extrinsic reward curve for agent with \textbf{IAMMIR}, \textbf{PPO}, \textbf{ICM}, \textbf{ECO} in Vizdoom during training. In $VerySparse$ and $Flytrap$ (extremely sparse extrinsic reward), \textbf{PPO} can't learn the policy to solve the task, but after adding the intrinsic reward, the agent can solve it, and our \textbf{IAMMIR} is the fastest to achieve stable task completion among the intrinsic reward algorithms, with the highest data efficiency. Besides, in $Dense$, where the agent is randomly spawned in 17 locations, our \textbf{IAMMIR} is still the fastest to achieve stable task completion, indicating that \textbf{IAMMIR} has a certain environment generalization.}
    \label{fig4}
\end{figure*}

\section{Experiment}

In this section, we verify the performance of the proposed method \textbf{IAMMIR} in a visual environment with sparse reward, a maze navigation scene ""Vizdoom" with the discrete action space. Training step of all environments is less than ten million steps. We compare our method \textbf{IAMMIR} with baseline \textbf{PPO}\cite{schulman2017proximal} and two intrinsic reward algorithms (\textbf{ICM}\cite{pathak2017curiosity}, \textbf{ECO}\cite{savinov2018episodic}).

\subsection{Environment and Setup}

Vizdoom \cite{kempka2016vizdoom} provides rich maze-like 3D environments. We test our method on two 3D navigation task scenes. One is VizdoomMyWayHome in Fig \ref{fig3} which contains 9 rooms. In this environment, the agent only accepts the image of the first-person view to decide the suitable discrete action. After Finding the armour, the agent receives a reward of +1, and the reward remains 0 during the rest time. An episode ends either when the agent finds the armour or when the agent has taken 2100 steps. This scene has two sub-scenarios: $VerySparse$, the agent will be spawned at the farthest position from the target armour in each episode to test the exploration efficiency;  $Dense$, the agent will be randomly spawned from 17 locations in each episode to test the scene generalizability. The other is VizdoomFlytrap inspired by flytrap in Fig \ref{fig3}, this environment is derived from Zhang's work \cite{zhang2019scheduled}. The maze consists of 4 rooms separated by V-shaped walls pointing inwards the rooms. The small exists of each room is located at the junction of the V-shape, extremely difficult to step into without a sequence of precise movements. Compared to VizdoomMyWayHome, each episode can take up to 10,000 steps. It is a more complex visual sparse reward scene than $VerySparse$. The action space size of all environments is 5, and A fixed action repeat 4 is applied across all environments. The input state is a grayscale-processed image of $4 \times 84 \times 84$ consecutive 4 moments. Meanwhile, so as to speed and efficiency, we run 8 environments in parallel.

\subsection{Network Setup and Hyperparameters}

The online and target encoders $f_{\theta}$ and $f_{\xi}$ both use the architecture from \textbf{ICM}. The online and target projectors $g_{\theta}$ and $g_{\xi}$ are 2-layer MLP with a batch normalization (BN) \cite{ioffe2015batch} and a ReLU non-linearities, which extract latent to 256 dimensions. The predictor $q_\theta$ is a 2-layer MLP with BN and ReLU. The PPO head $p_\theta$ contains the actor module and the critic module, both of them are a 2-layer MLP with ReLU.

We train the online network parameters $\theta$ using stochastic gradient optimization with Adam \cite{kingma2014adam}, where the learning rate is set to $2.5 \times 10^{-4}$ and mini batch size is 256. The target network parameters $\xi$ are updated as an exponential moving average of $\theta$ with momentum $\tau=0.001$. The factor between intrinsic and extrinsic rewards is $\beta=0.1$. The coefficient $\alpha$ in $\mathcal{L}^{total}$ is to balance the influence between $\mathcal{L}^{PPO}$ and $\mathcal{L}^{IAFDR}$, we set it to 0.2.

\subsection{Navigation with Sparse Extrinsic Rewards}

The average extrinsic reward curve in the period of training for the experiment is shown in Fig \ref{fig4}. We can draw the following conclusions. First, our method \textbf{IAMMIR} can stably achieve the goal in all visual environments with sparse extrinsic rewards. Second, our method consistently beats the baseline PPO and exceeds the previous intrinsic algorithms \textbf{ICM} and \textbf{ECO} in all environments. Compared to \textbf{ICM}, our method is at least 7 times and 6 times faster than it in reaching 100\% success rate in $VerySparse$ and $Flytrap$, respectively. Compared to the previous state-of-the-art \textbf{ECO}, our method is at least 2 times faster than it in reaching 100\% success rate in $VerySparse$ and $Flytrap$. As far as we know, our \textbf{IAMMIR} is the state-of-the-art performance in Vizdoom navigation task. Finally, the result in $Dense$ shows that \textbf{IAMMIR} has the great scene generalization, and also gets the state-of-the-art performance in sample efficiency.

\begin{figure}[h]
    \centering
    \subfigure[$VerySparse$]{
    \begin{minipage}[t]{1.\linewidth}
    \centering
    \includegraphics[scale=0.035]{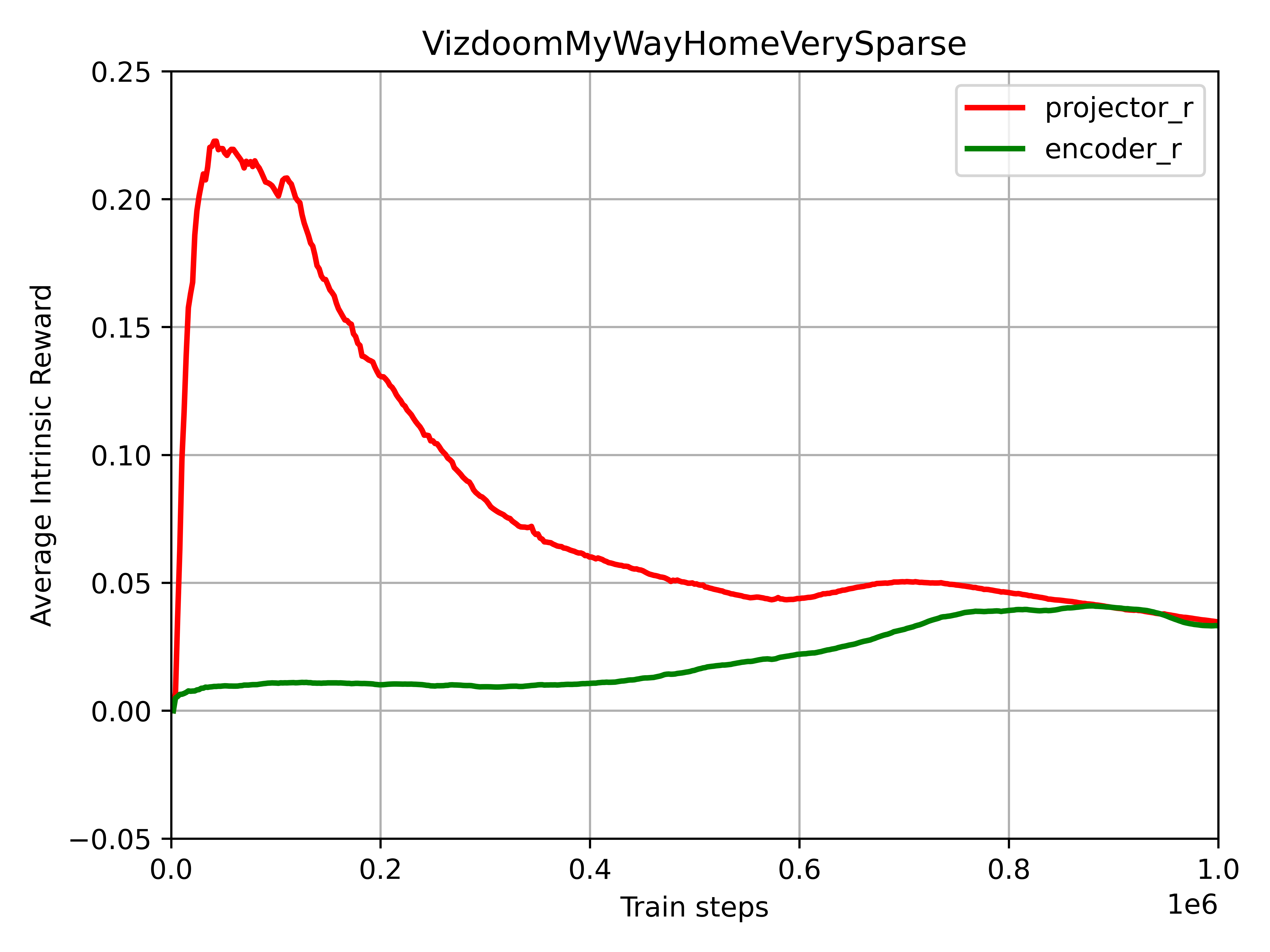}
    \end{minipage}%
    }%
    
    \subfigure[$Flytrap$]{
    \begin{minipage}[t]{1.\linewidth}
    \centering
    \includegraphics[scale=0.035]{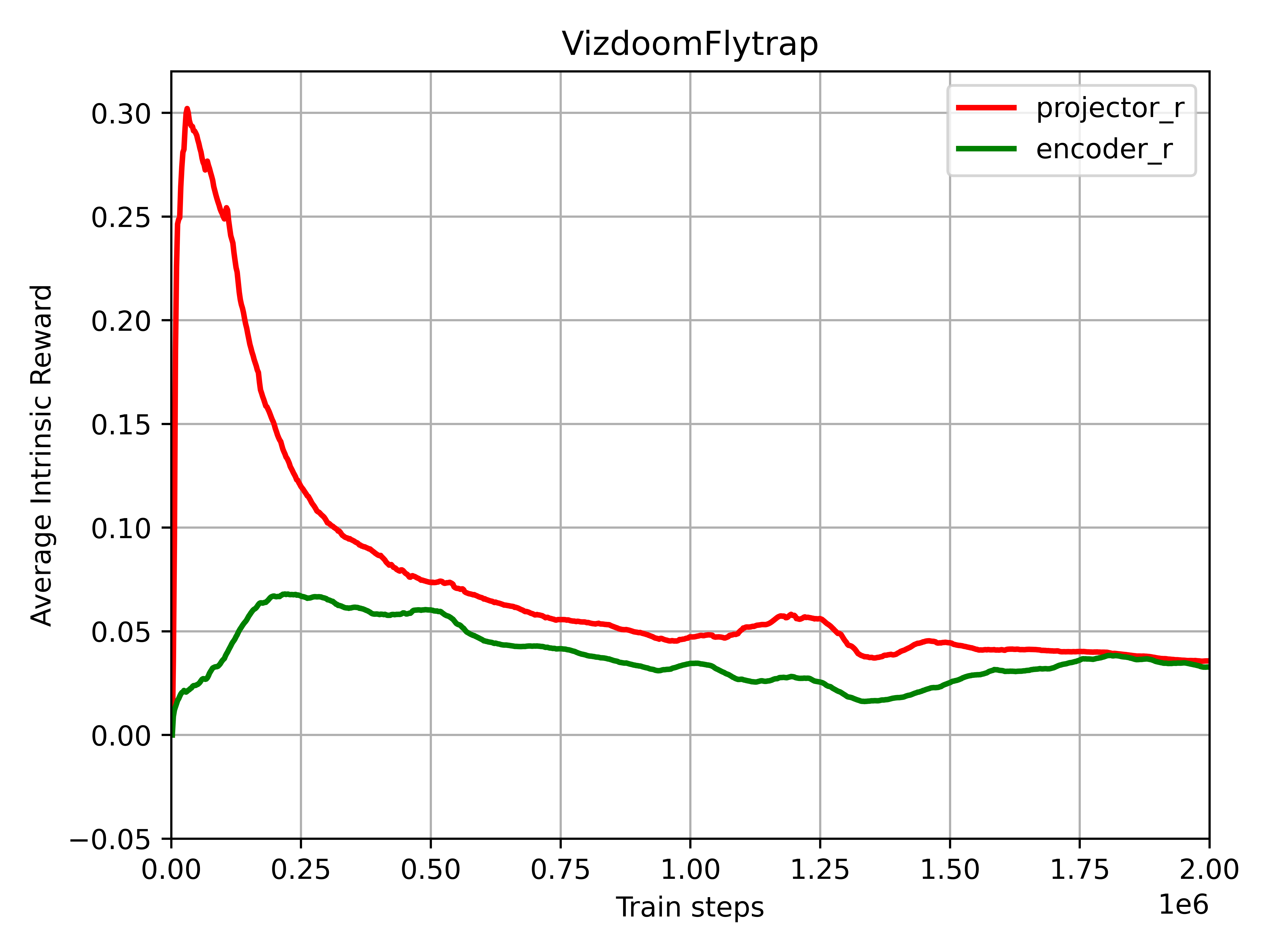}
    \end{minipage}%
    }%
    \caption{the average intrinsic reward curve for agent with \textbf{IAMMIR} in Vizdoom during training. The red is \textbf{MMIR} with the output of projector $g$, it will decrease gradually along with train steps, indicating the process of agent's familiarity of environment. The green is \textbf{MMIR} with the output of encoder $f$, it doesn't show a downward trend, on the contrary it is gradually rising somewhere.}
    \label{fig5}
\end{figure}

\subsection{Intrinsic Reward in New RL Objective}
\label{sbusection:4.4}
In the new RL objective, the intrinsic reward will affect the learning of the encoder layer $f_\theta$. As discussed in previous article \cite{raileanu2020ride}, if the representation used in the intrinsic reward is the same as the representation of the policy, the agent can artificially maximize its intrinsic reward by constructing state representations with large distances among themselves, without grounding them in environment states. As shown in Fig \ref{fig5}, we find there will be such problem in both $VerySparse$ and $Flytrap$, if the output of the encoder is used directly in the \textbf{MMIR}. But in our work, we turn to the output of the projector $g_{\theta}$ to compute the \textbf{MMIR}. It is found that the intrinsic reward with the representation of the projector satisfies the basic hypothesis that state familiarity value will decrease gradually along with the training time. At the same time, in theory, the intrinsic reward in the new RL objective will not affect the parameters of projector $g_\theta$, which can accurately characterise the environment states.

\subsection{Ablation Study}
\begin{figure}[h]
    \centering
    \includegraphics[scale=0.035]{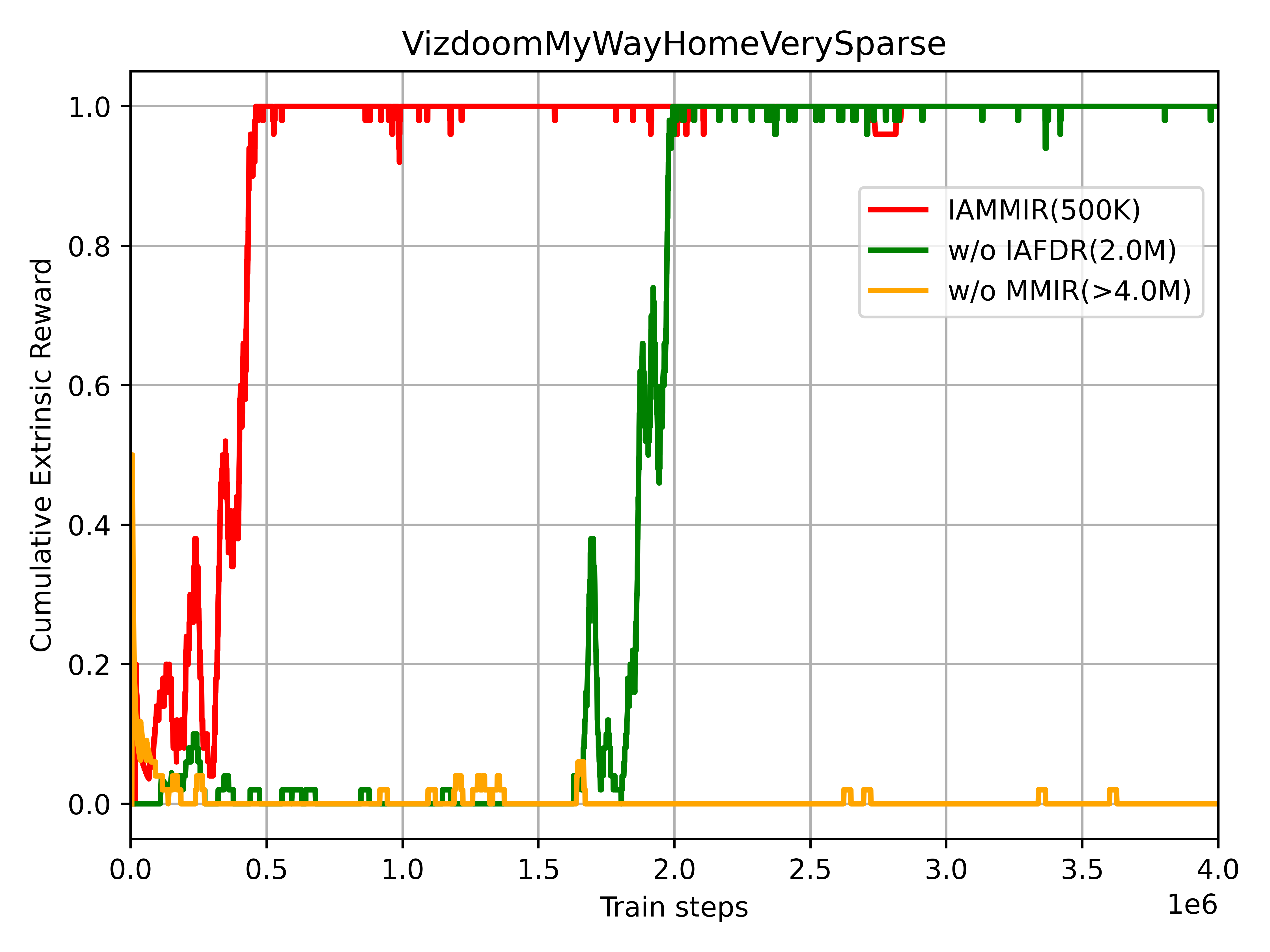}
    \caption{the cumulative extrinsic rewards curve for agent with \textbf{IAMMIR}, without \textbf{MMIR} and without \textbf{IAFDR} during training. We compare \textbf{IAMMIR} (red) to two ablations (green and yellow).}
    \label{fig6}
\end{figure}

To investigate the contributions of the components within the proposed method, especially, the \textbf{IAFDR} and \textbf{MMIR}. We conducted an ablation analysis by the agent without \textbf{IAFDR} or without \textbf{MMIR} in $VerySparse$. As shown in Fig \ref{fig6}, the \textbf{MMIR} module is the most crucial component of \textbf{IAMMIR}, removing it caused failure in $VerySparse$ task. Unsurprisingly, the removal of \textbf{IAFDR} also causes a large drop in learning speed. The same is true, for the sparse reward visual scenes, a good representation needs to be driven by diverse data and dense reward feedback, which comes from sufficient exploration. Meanwhile, sufficient exploration requires a good representation to distinguish between the visited states and novel states. Therefore, it is a very worthwhile solution for the sparse reward visual scenes to combine the visual representation learning with the intrinsic reward algorithms.

\section{Conclusion}

To address the problem of the sparse reward visual scenes, we decompose it into two sub-problems: the visual representation and the sparse reward. For the two sub-problems, a novel method \textbf{IAMMIR} is propoesed. Specifically, a sufficient state representation is acquired by the self-supervised representation learning, on the basis of which a novel momentum memory intrinsic reward \textbf{MMIR} is designed. We conduct experiments in three visual navigation scenes in Vizdoom and show that our \textbf{IAMMIR} helps greatly in improving the exploration efficiency and achieves the state-of-the-art performance. We believe that \textbf{IAMMIR} is a superior framework for sparse reward visual scenes. Our next step is to apply our proposed framework to the environment with continuous action space like manipulation, and explore more applications.


\bibliography{ref}

\begin{thebibliography}{36}
\providecommand{\natexlab}[1]{#1}
\providecommand{\url}[1]{\texttt{#1}}
\expandafter\ifx\csname urlstyle\endcsname\relax
  \providecommand{\doi}[1]{doi: #1}\else
  \providecommand{\doi}{doi: \begingroup \urlstyle{rm}\Url}\fi

\bibitem[Badia et~al.(2020)Badia, Piot, Kapturowski, Sprechmann, Vitvitskyi,
  Guo, and Blundell]{badia2020agent57}
Badia, A.~P., Piot, B., Kapturowski, S., Sprechmann, P., Vitvitskyi, A., Guo,
  Z.~D., and Blundell, C.
\newblock Agent57: Outperforming the atari human benchmark.
\newblock In \emph{International Conference on Machine Learning}, pp.\
  507--517. PMLR, 2020.

\bibitem[Bellemare et~al.(2016)Bellemare, Srinivasan, Ostrovski, Schaul,
  Saxton, and Munos]{bellemare2016unifying}
Bellemare, M., Srinivasan, S., Ostrovski, G., Schaul, T., Saxton, D., and
  Munos, R.
\newblock Unifying count-based exploration and intrinsic motivation.
\newblock \emph{Advances in neural information processing systems}, 29, 2016.

\bibitem[Bellman(1957)]{bellman1957markovian}
Bellman, R.
\newblock A markovian decision process.
\newblock \emph{Journal of mathematics and mechanics}, pp.\  679--684, 1957.

\bibitem[Berner et~al.(2019)Berner, Brockman, Chan, Cheung, Dkebiak, Dennison,
  Farhi, Fischer, Hashme, Hesse, et~al.]{berner2019dota}
Berner, C., Brockman, G., Chan, B., Cheung, V., Dkebiak, P., Dennison, C.,
  Farhi, D., Fischer, Q., Hashme, S., Hesse, C., et~al.
\newblock Dota 2 with large scale deep reinforcement learning.
\newblock \emph{arXiv preprint arXiv:1912.06680}, 2019.

\bibitem[Burda et~al.(2018{\natexlab{a}})Burda, Edwards, Pathak, Storkey,
  Darrell, and Efros]{burda2018large}
Burda, Y., Edwards, H., Pathak, D., Storkey, A., Darrell, T., and Efros, A.~A.
\newblock Large-scale study of curiosity-driven learning.
\newblock \emph{arXiv preprint arXiv:1808.04355}, 2018{\natexlab{a}}.

\bibitem[Burda et~al.(2018{\natexlab{b}})Burda, Edwards, Storkey, and
  Klimov]{burda2018exploration}
Burda, Y., Edwards, H., Storkey, A., and Klimov, O.
\newblock Exploration by random network distillation.
\newblock \emph{arXiv preprint arXiv:1810.12894}, 2018{\natexlab{b}}.

\bibitem[Caron et~al.(2020)Caron, Misra, Mairal, Goyal, Bojanowski, and
  Joulin]{caron2020unsupervised}
Caron, M., Misra, I., Mairal, J., Goyal, P., Bojanowski, P., and Joulin, A.
\newblock Unsupervised learning of visual features by contrasting cluster
  assignments.
\newblock \emph{Advances in Neural Information Processing Systems},
  33:\penalty0 9912--9924, 2020.

\bibitem[Chen et~al.(2020)Chen, Kornblith, Norouzi, and Hinton]{chen2020simple}
Chen, T., Kornblith, S., Norouzi, M., and Hinton, G.
\newblock A simple framework for contrastive learning of visual
  representations.
\newblock In \emph{International conference on machine learning}, pp.\
  1597--1607. PMLR, 2020.

\bibitem[Grill et~al.(2020)Grill, Strub, Altch{\'e}, Tallec, Richemond,
  Buchatskaya, Doersch, Avila~Pires, Guo, Gheshlaghi~Azar,
  et~al.]{grill2020bootstrap}
Grill, J.-B., Strub, F., Altch{\'e}, F., Tallec, C., Richemond, P.,
  Buchatskaya, E., Doersch, C., Avila~Pires, B., Guo, Z., Gheshlaghi~Azar, M.,
  et~al.
\newblock Bootstrap your own latent-a new approach to self-supervised learning.
\newblock \emph{Advances in Neural Information Processing Systems},
  33:\penalty0 21271--21284, 2020.

\bibitem[He et~al.(2020)He, Fan, Wu, Xie, and Girshick]{he2020momentum}
He, K., Fan, H., Wu, Y., Xie, S., and Girshick, R.
\newblock Momentum contrast for unsupervised visual representation learning.
\newblock In \emph{Proceedings of the IEEE/CVF conference on computer vision
  and pattern recognition}, pp.\  9729--9738, 2020.

\bibitem[Ioffe \& Szegedy(2015)Ioffe and Szegedy]{ioffe2015batch}
Ioffe, S. and Szegedy, C.
\newblock Batch normalization: Accelerating deep network training by reducing
  internal covariate shift.
\newblock In \emph{International conference on machine learning}, pp.\
  448--456. PMLR, 2015.

\bibitem[Kempka et~al.(2016)Kempka, Wydmuch, Runc, Toczek, and
  Ja{\'s}kowski]{kempka2016vizdoom}
Kempka, M., Wydmuch, M., Runc, G., Toczek, J., and Ja{\'s}kowski, W.
\newblock Vizdoom: A doom-based ai research platform for visual reinforcement
  learning.
\newblock In \emph{2016 IEEE conference on computational intelligence and games
  (CIG)}, pp.\  1--8. IEEE, 2016.

\bibitem[Kingma \& Ba(2014)Kingma and Ba]{kingma2014adam}
Kingma, D.~P. and Ba, J.
\newblock Adam: A method for stochastic optimization.
\newblock \emph{arXiv preprint arXiv:1412.6980}, 2014.

\bibitem[Kostrikov et~al.(2020)Kostrikov, Yarats, and
  Fergus]{kostrikov2020image}
Kostrikov, I., Yarats, D., and Fergus, R.
\newblock Image augmentation is all you need: Regularizing deep reinforcement
  learning from pixels.
\newblock \emph{arXiv preprint arXiv:2004.13649}, 2020.

\bibitem[Laskin et~al.(2020)Laskin, Lee, Stooke, Pinto, Abbeel, and
  Srinivas]{laskin2020reinforcement}
Laskin, M., Lee, K., Stooke, A., Pinto, L., Abbeel, P., and Srinivas, A.
\newblock Reinforcement learning with augmented data.
\newblock \emph{Advances in Neural Information Processing Systems},
  33:\penalty0 19884--19895, 2020.

\bibitem[Mnih et~al.(2013)Mnih, Kavukcuoglu, Silver, Graves, Antonoglou,
  Wierstra, and Riedmiller]{mnih2013playing}
Mnih, V., Kavukcuoglu, K., Silver, D., Graves, A., Antonoglou, I., Wierstra,
  D., and Riedmiller, M.
\newblock Playing atari with deep reinforcement learning.
\newblock \emph{arXiv preprint arXiv:1312.5602}, 2013.

\bibitem[Ostrovski et~al.(2017)Ostrovski, Bellemare, Oord, and
  Munos]{ostrovski2017count}
Ostrovski, G., Bellemare, M.~G., Oord, A., and Munos, R.
\newblock Count-based exploration with neural density models.
\newblock In \emph{International conference on machine learning}, pp.\
  2721--2730. PMLR, 2017.

\bibitem[Oudeyer et~al.(2007)Oudeyer, Kaplan, and Hafner]{oudeyer2007intrinsic}
Oudeyer, P.-Y., Kaplan, F., and Hafner, V.~V.
\newblock Intrinsic motivation systems for autonomous mental development.
\newblock \emph{IEEE transactions on evolutionary computation}, 11\penalty0
  (2):\penalty0 265--286, 2007.

\bibitem[Pathak et~al.(2017)Pathak, Agrawal, Efros, and
  Darrell]{pathak2017curiosity}
Pathak, D., Agrawal, P., Efros, A.~A., and Darrell, T.
\newblock Curiosity-driven exploration by self-supervised prediction.
\newblock In \emph{International conference on machine learning}, pp.\
  2778--2787. PMLR, 2017.

\bibitem[Pathak et~al.(2019)Pathak, Gandhi, and Gupta]{pathak2019self}
Pathak, D., Gandhi, D., and Gupta, A.
\newblock Self-supervised exploration via disagreement.
\newblock In \emph{International conference on machine learning}, pp.\
  5062--5071. PMLR, 2019.

\bibitem[Raileanu \& Rockt{\"a}schel(2020)Raileanu and
  Rockt{\"a}schel]{raileanu2020ride}
Raileanu, R. and Rockt{\"a}schel, T.
\newblock Ride: Rewarding impact-driven exploration for procedurally-generated
  environments.
\newblock \emph{arXiv preprint arXiv:2002.12292}, 2020.

\bibitem[Riba et~al.(2020)Riba, Mishkin, Ponsa, Rublee, and
  Bradski]{riba2020kornia}
Riba, E., Mishkin, D., Ponsa, D., Rublee, E., and Bradski, G.
\newblock Kornia: an open source differentiable computer vision library for
  pytorch.
\newblock In \emph{Proceedings of the IEEE/CVF Winter Conference on
  Applications of Computer Vision}, pp.\  3674--3683, 2020.

\bibitem[Savinov et~al.(2018)Savinov, Raichuk, Marinier, Vincent, Pollefeys,
  Lillicrap, and Gelly]{savinov2018episodic}
Savinov, N., Raichuk, A., Marinier, R., Vincent, D., Pollefeys, M., Lillicrap,
  T., and Gelly, S.
\newblock Episodic curiosity through reachability.
\newblock \emph{arXiv preprint arXiv:1810.02274}, 2018.

\bibitem[Schmidhuber(2010)]{schmidhuber2010formal}
Schmidhuber, J.
\newblock Formal theory of creativity, fun, and intrinsic motivation
  (1990--2010).
\newblock \emph{IEEE transactions on autonomous mental development}, 2\penalty0
  (3):\penalty0 230--247, 2010.

\bibitem[Schulman et~al.(2015)Schulman, Moritz, Levine, Jordan, and
  Abbeel]{schulman2015high}
Schulman, J., Moritz, P., Levine, S., Jordan, M., and Abbeel, P.
\newblock High-dimensional continuous control using generalized advantage
  estimation.
\newblock \emph{arXiv preprint arXiv:1506.02438}, 2015.

\bibitem[Schulman et~al.(2017)Schulman, Wolski, Dhariwal, Radford, and
  Klimov]{schulman2017proximal}
Schulman, J., Wolski, F., Dhariwal, P., Radford, A., and Klimov, O.
\newblock Proximal policy optimization algorithms.
\newblock \emph{arXiv preprint arXiv:1707.06347}, 2017.

\bibitem[Schwarzer et~al.(2020)Schwarzer, Anand, Goel, Hjelm, Courville, and
  Bachman]{schwarzer2020data}
Schwarzer, M., Anand, A., Goel, R., Hjelm, R.~D., Courville, A., and Bachman,
  P.
\newblock Data-efficient reinforcement learning with self-predictive
  representations.
\newblock \emph{arXiv preprint arXiv:2007.05929}, 2020.

\bibitem[Sermanet et~al.(2018)Sermanet, Lynch, Chebotar, Hsu, Jang, Schaal,
  Levine, and Brain]{sermanet2018time}
Sermanet, P., Lynch, C., Chebotar, Y., Hsu, J., Jang, E., Schaal, S., Levine,
  S., and Brain, G.
\newblock Time-contrastive networks: Self-supervised learning from video.
\newblock In \emph{2018 IEEE international conference on robotics and
  automation (ICRA)}, pp.\  1134--1141. IEEE, 2018.

\bibitem[Srinivas et~al.(2020)Srinivas, Laskin, and Abbeel]{srinivas2020curl}
Srinivas, A., Laskin, M., and Abbeel, P.
\newblock Curl: Contrastive unsupervised representations for reinforcement
  learning.
\newblock \emph{arXiv preprint arXiv:2004.04136}, 2020.

\bibitem[Tai et~al.(2017)Tai, Paolo, and Liu]{tai2017virtual}
Tai, L., Paolo, G., and Liu, M.
\newblock Virtual-to-real deep reinforcement learning: Continuous control of
  mobile robots for mapless navigation.
\newblock In \emph{2017 IEEE/RSJ International Conference on Intelligent Robots
  and Systems (IROS)}, pp.\  31--36. IEEE, 2017.

\bibitem[Vinyals et~al.(2019)Vinyals, Babuschkin, Czarnecki, Mathieu, Dudzik,
  Chung, Choi, Powell, Ewalds, Georgiev, et~al.]{vinyals2019grandmaster}
Vinyals, O., Babuschkin, I., Czarnecki, W.~M., Mathieu, M., Dudzik, A., Chung,
  J., Choi, D.~H., Powell, R., Ewalds, T., Georgiev, P., et~al.
\newblock Grandmaster level in starcraft ii using multi-agent reinforcement
  learning.
\newblock \emph{Nature}, 575\penalty0 (7782):\penalty0 350--354, 2019.

\bibitem[Vondrick et~al.(2018)Vondrick, Shrivastava, Fathi, Guadarrama, and
  Murphy]{vondrick2018tracking}
Vondrick, C., Shrivastava, A., Fathi, A., Guadarrama, S., and Murphy, K.
\newblock Tracking emerges by colorizing videos.
\newblock In \emph{Proceedings of the European conference on computer vision
  (ECCV)}, pp.\  391--408, 2018.

\bibitem[Wang et~al.(2019)Wang, Jabri, and Efros]{wang2019learning}
Wang, X., Jabri, A., and Efros, A.~A.
\newblock Learning correspondence from the cycle-consistency of time.
\newblock In \emph{Proceedings of the IEEE/CVF Conference on Computer Vision
  and Pattern Recognition}, pp.\  2566--2576, 2019.

\bibitem[Yarats et~al.(2021)Yarats, Fergus, Lazaric, and
  Pinto]{yarats2021reinforcement}
Yarats, D., Fergus, R., Lazaric, A., and Pinto, L.
\newblock Reinforcement learning with prototypical representations.
\newblock In \emph{International Conference on Machine Learning}, pp.\
  11920--11931. PMLR, 2021.

\bibitem[Zhang et~al.(2019)Zhang, Wetzel, Dorka, Boedecker, and
  Burgard]{zhang2019scheduled}
Zhang, J., Wetzel, N., Dorka, N., Boedecker, J., and Burgard, W.
\newblock Scheduled intrinsic drive: A hierarchical take on intrinsically
  motivated exploration.
\newblock \emph{arXiv preprint arXiv:1903.07400}, 2019.

\bibitem[Zhelo et~al.(2018)Zhelo, Zhang, Tai, Liu, and
  Burgard]{zhelo2018curiosity}
Zhelo, O., Zhang, J., Tai, L., Liu, M., and Burgard, W.
\newblock Curiosity-driven exploration for mapless navigation with deep
  reinforcement learning.
\newblock \emph{arXiv preprint arXiv:1804.00456}, 2018.

\end{thebibliography}
\bibliographystyle{icml2021}

\end{document}